\documentclass[runningheads]{llncs}
\usepackage{graphicx}
\usepackage{comment}
\usepackage{bbm}
\usepackage{amsmath,amssymb} 
\usepackage{color}

\usepackage{caption}
\usepackage{ulem}
\usepackage{xcolor}
\usepackage{float}
\usepackage[pagebackref=true,breaklinks=true,letterpaper=true,colorlinks,bookmarks=false]{hyperref}

\begin{document}
\pagestyle{headings}
\mainmatter
\def\ECCVSubNumber{3136}  

\title{Online Ensemble Model Compression using Knowledge Distillation } 

\titlerunning{Online Ensemble Model Compression}
%
\author{Devesh Walawalkar \and
Zhiqiang Shen \and
Marios Savvides}
\authorrunning{D. Walawalkar et al.}
%
\institute{Carnegie Mellon University, Pittsburgh PA 15213, USA \\
\email{devwalkar64@gmail.com \{zhiqians,marioss\}@andrew.cmu.edu}}
\maketitle

\begin{abstract}
This paper presents a novel knowledge distillation based model compression framework consisting of a student ensemble. It enables distillation of simultaneously learnt ensemble knowledge onto each of the compressed student models. Each model learns unique representations from the data distribution due to its distinct architecture. This helps the ensemble generalize better by combining every model's knowledge. The distilled students and ensemble teacher are trained simultaneously without requiring any pretrained weights. Moreover, our proposed method can deliver multi-compressed students with single training, which is efficient and flexible for different scenarios. We provide comprehensive experiments using state-of-the-art classification models to validate our framework's effectiveness. Notably, using our framework a 97\% compressed ResNet110 student model managed to produce a 10.64\% relative accuracy gain over its individual baseline training on CIFAR100 dataset. Similarly a 95\% compressed DenseNet-BC (k=12) model managed a 8.17\% relative accuracy gain.    

\end{abstract}

\section{Introduction}
Deep Learning based neural networks have provided tremendous improvements over the past decade in various domains of Computer Vision. These include Image Classification \cite{krizhevsky2012imagenet,he2016deep,huang2017densely,tan2019efficientnet}, Object Detection \cite{ren2015faster,cai2018cascade,liu2016ssd,redmon2016you,duan2019centernet}, Semantic Segmentation \cite{he2017mask,chen2017rethinking,zhang2018exfuse,chen2018encoder,zhong2019squeeze} among others. The drawbacks of these methods however include the fact that a large amount of computational resources are required to achieve state-of-the-art accuracy. A trend started setting in where constructing deeper and wider models provided better accuracy at the cost of considerable resource utilization \cite{simonyan2014very,szegedy2015going,liu2019cbnet}. The difference in resource utilization is considerable compared to traditional computer vision techniques. To alleviate this gap, model compression techniques started being developed to reduce these large computational requirements. These techniques can broadly be classified into four types \cite{cheng2017survey} i.e. Parameter Pruning \cite{hanson1989comparing,srinivas2015data,chen2015compressing,han2015learning}, Low Rank Factorization \cite{cheng2015exploration,rakhuba2015fast,sindhwani2015structured}, Transferred Convolutional Filters \cite{cohen2016group,shang2016understanding,li2016multi,dieleman2016exploiting} and Knowledge Distillation methods  \cite{hinton2015distilling,kim2016sequence,adriana2015fitnets,furlanello2018born,koratana2019lit,zhu2018knowledge}. Each of these types was able to provide impressive computational reductions while simultaneously managing to keep the accuracy degradation to a minimum.\par

\begin{figure}[t]
    \centering
    \includegraphics[width=\linewidth]{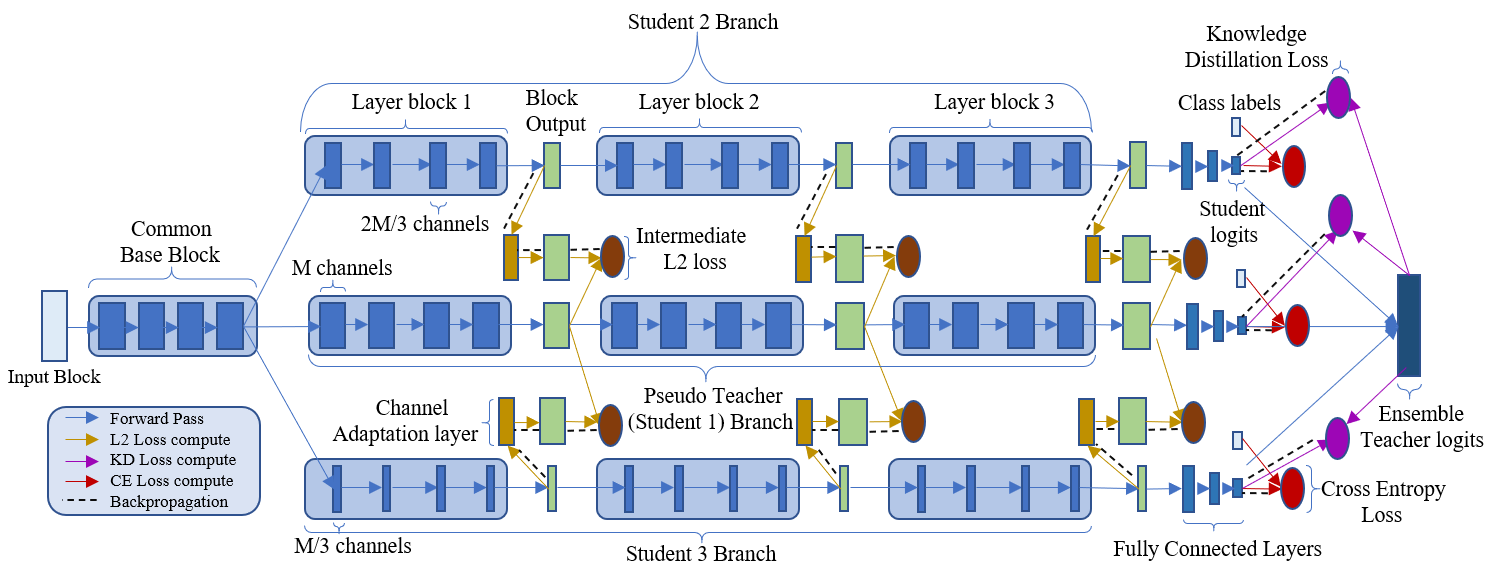}
    \caption{Overview of our \textit{model compression framework} for a \textit{3 student ensemble}. Each student is composed of the base block and one of the network branches on top of it. The original model is the first student in the ensemble, termed as \textit{pseudo teacher} due to its simultaneous knowledge transfer capability and training from scratch properties. The \textit{ensemble teacher} is a weighted combination of all student's output logits. Each student is divided into four blocks such that each block incorporates approximately the same number of layers. The layer channels for the compressed student branches are reduced by a specific ratio with respect to the original model. For a 3 student ensemble, the layer channels assigned are M, 2M/3 and M/3, where M is the original layer channel count. \textit{Channel adaptation layers} help map the compressed student's block output channels to the pseudo teacher's output channels in order to compute an intermediate feature representation loss} 
    \label{fig:block_diagram}
\end{figure}

Knowledge Distillation (KD) in particular has provided great model compression capabilities using a novel teacher-student model concept \cite{hinton2015distilling}. Here a teacher, the original model trained for a specific task is used to teach a compressed or replicated version of itself referred to as student. The student is encouraged to mimic the teacher output distribution, which helps the student generalize much better and in certain cases leads to the student performing better than the teacher itself. Drawbacks of these methods include the fact that a pre-trained model is required to distill knowledge onto a student model. To solve this, simultaneous distillation methods \cite{anil2018large,zhu2018knowledge} were developed wherein multiple students were being trained with an ensemble teacher learning on the fly with the students itself in an ensemble training scheme. These methods however primarily focused on replicating the teacher multiple times, causing the ensemble to gain sub-optimal generalized knowledge due to model architecture redundancy. Also in parallel, methods \cite{koratana2019lit,shen2019meal,adriana2015fitnets} were developed which focused on distilling not only the output knowledge but also the intermediate representations of teacher onto the student to have more effective knowledge transfer. These techniques are efficient, however still suffering from the drawback of requiring teacher model pre-training in addition to the student training.\par
In this paper, we present a novel model compression and ensemble training framework which improves on all aforementioned drawbacks and provides a new perspective for ensemble model compression techniques. We present a framework which enables multiple student model training using knowledge transfer from an \textit{ensemble teacher}. Each of these student models represents a version of the original model compressed to different degrees. Knowledge is distilled onto each of the compressed student through the ensemble teacher and also through intermediate representations from a \textit{pseudo teacher}. The original model is the first student in our ensemble, termed as pseudo teacher due to its simultaneous knowledge transfer capability and training from scratch property. Moreover this framework simultaneously provides multiple compressed models having different computational budget with each having benefited from every other model's training. Our framework facilitates the choice of selecting a student that fits the resource budget for a particular use case with the knowledge that every student provides decent comparable performance to the original model. Also, the ensemble training significantly reduces the training time of all student combined, compared to when they are trained individually. We specifically focus on image classification task while providing extensive experimentation on popular image classification models and bench-marked datasets.\par

\noindent{Our} contributions from this paper can be thus summarized as follows:
\begin{enumerate}
    \item  We present a novel ensemble model compression framework based on knowledge distillation that can generate multiple compressed networks simultaneously with a single training run and is completely model agnostic.
    \item We provide extensive experiments on popular classification models on standard datasets with corresponding baseline comparisons of each individual student training to provide evidence of our framework's effectiveness.
    \item We provide hyper parameter ablation studies which help provide insights into the effective ensemble knowledge distillation provided by our framework.
\end{enumerate}

\section{Related Works}
\subsection{Model Compression using Knowledge Distillation}
Hinton et al. \cite{hinton2015distilling} introduced the concept of distilling knowledge from a larger teacher model onto a smaller compressed student model. Mathematically this meant training the student on softened teacher output distribution in addition to the traditional cross entropy with dataset labels. The paper argued that the teacher distribution provided much richer information about an image compared to just one hot labels. For e.g. consider a classification task of differentiating between various breeds of dogs. The output distribution of a higher capability teacher provides the student with the information of how alike one breed of dogs looks to the other. This helps the student learn more generalized features of each dog breed compared to providing just one hot labels which fails to provide any comparative knowledge. Also, in process of trying to mimic the distribution of a much deeper teacher model the student tries to find a compact series of transformation that tries to mimic the teacher's larger series of transformations. This inherently helps the student negate the accuracy loss due to compression. Impressively, in certain cases the student manages to outperform its teacher due to this superior generalization training. Further works extended this concept by using internal feature representations \cite{adriana2015fitnets}, adversarial learning \cite{shen2019meal} and inner product between feature maps \cite{yim2017gift}.

\subsection{Online Knowledge Distillation}
The single practical drawback of Knowledge Distillation is the fact that a pre-trained model is required to play the role of a teacher. This entails multiple sequential training runs on the same dataset. Anil et al. \cite{anil2018large} came up with a method to train the student in parallel with the teacher termed as codistillation. Here the student is an exact replica of the teacher and roles of teacher and student were continuously interchanged between the models during training with one training the other iteratively. The primary distinguishing property between the models was their distinct parameter initialization. This enabled each model to learn unique features which were then distilled from one to the other as training progressed.\par
Zhang et al. \cite{zhang2018deep} employed a unique KD loss, by using KL divergence loss between two model output distributions to penalize the differences between them. Each model's training involved a combination loss of KL divergence loss with other model's distribution and traditional cross entropy loss. Both models acting as teacher and student simultaneously were trained jointly in an online fashion.\par
Lan et al. \cite{zhu2018knowledge} extended this online KD concept by having multiple replicas of a given model in a multi branch architecture fashion. Multi branch architecture designs became popular with the image classification models like Resnet \cite{he2016deep}, Inception \cite{szegedy2015going,szegedy2016rethinking} and ResNext \cite{xie2017aggregated}. In this paper, the multiple replicated models have a common base of block of layers with each model represented as a branch on top of this base with subsequent layer blocks from the original model architecture right until the final fully connected layers. The teacher in this concept was the combined output of all the student models in the ensemble. Each student model learnt from the ensemble joint knowledge represented by the teacher outputs. Our paper builds on this core concept however is fundamentally different as we incorporate compressed student branches, more efficient training procedure, incorporate intermediate representation distillation in addition to the final output distillation among others. 

\subsection{Intermediate Representation Knowledge Distillation}
A separate branch of Knowledge Distillation focuses on training the student to mimic the intermediate representations obtained in form of feature maps from certain intermediate layer blocks within the teacher. This provides a more stricter learning regime for the student who has to focus not only on the final teacher output distribution but also on its intermediate layer feature maps. Romero et al. \cite{adriana2015fitnets} provide one of the preliminary works in this direction by distilling a single intermediate layer's knowledge onto the student which they term as providing hint to the student. Mathematically, hint training involves minimizing a combination of L2 loss between the features maps at an immediate layer of the two models and the regular KD (KL divergence loss) between the output distributions.\par
Koratana et al. \cite{koratana2019lit} extend this concept by comparing feature maps at not just one but multiple intermediate locations within the model which can be related to as multiple hint training. This method however again requires a pre-trained teacher model which might be time consuming and compute expensive for certain real world scenarios. Our method incorporates multiple hint training for all student models with respect to pseudo teacher, which is also the first student in the ensemble. A network's depth measures its function modeling capacity in general. Koratana et al. \cite{koratana2019lit} compress the model by removing blocks of layers from the network which severely affects the network depth. Our work incorporates a more robust compression logic where the number of channels in every student layer are simply reduced by a certain percent, thus preserving the model depth.         

\section{Methodology}
An overview of our ensemble compression framework is presented in Figure \ref{fig:block_diagram}, which can be split up into three major sections for the ease of understanding. We would be going into their details in the following sections.

\subsection{Ensemble Student Model Compression}
First, the entire architecture of a given neural network is broken down into a series of layer blocks, ideally into four blocks. The first block is designated as a common base block and the rest of the blocks are replicated in parallel to create branches as shown in Figure \ref{fig:block_diagram}. A single student model can be viewed as a series of base block and one of the branches on top of it. As previously mentioned, the original network is designated as the first student model, also termed as pseudo teacher. For every successive student branch the number of channels in each layer of its blocks is reduced by a certain ratio with respect to the pseudo teacher. This ratio becomes higher for every new student branch created. For example, for a four student ensemble and $C$ being number of channels in pseudo teacher, the channels in other three students are assigned to be 0.75$C$, 0.5$C$ and 0.25$C$. The students are compressed versions of the original model to varying degrees, which still manage to maintain the original network depth. The channels in the common base block are kept the same as original whose main purpose is to provide constant low level features to all the student branches.\par
The output logits from all the student models are averaged together to create the ensemble teacher logits. This ensemble teacher output distribution represents the joint knowledge of the ensemble. During inference stage, any of the individual student models can be selected from the ensemble depending on the computational hardware constraints. In case of lenient constraints, the entire ensemble can be used with the ensemble teacher providing inference based on the learnt ensemble knowledge. From our studies we find that having 5 students (inclusive of pseudo teacher) provides an optimal trade off between training time and effective model compression. Table \ref{tab:size_comparison} provides an overview of compressed student model sizes and their CIFAR10 \cite{krizhevsky2009learning} trained accuracies for a five student ensemble based on various classification model architectures.

\begin{table}[t]
    \centering
    \setlength{\tabcolsep}{2pt}
    \captionof{table}{ Model size and test set accuracy comparison of every student in a five student ensemble, with their percent relative size in respect to the original model and their CIFAR10 test accuracies achieved using our ensemble framework.}
    \resizebox{0.9\textwidth}{!}{%
    \begin{tabular}{c|c|c|c|c|c|c|c|c|c|c}
    \hline
    \bf Classification & \multicolumn{10}{c}{\bf Student Model size and accuracy (\%)}   \\\cline{2-11}
    \bf    Model   & \multicolumn{2}{c}{First} & \multicolumn{2}{c}{Second}  & \multicolumn{2}{c}{Third} & \multicolumn{2}{c}{Fourth} & \multicolumn{2}{c}{Fifth} \\\cline{2-11}
    & Size & Accuracy & Size & Accuracy & Size & Accuracy & Size & Accuracy & Size & Accuracy\\\hline
    
    ResNet20 \cite{he2016deep} &100.0 &92.13 &62.95&92.18 &35.61&91.78 &15.50& 91.45 &3.95 & 91.03  \\\hline
    ResNet33 \cite{he2016deep} &100.0 & 92.25 &62.87 & 92.76 &35.43 & 92.45 &15.33 & 92.11 &3.8 & 91.78  \\\hline
    ResNet44 \cite{he2016deep} &100.0 & 93.45 &62.84 & 93.29 &35.36 & 93.11 &15.25 & 92.89 &3.74 & 92.56  \\\hline
    ResNet110 \cite{he2016deep} &100.0 & 94.24 &62.79 & 94.18 &35.26 & 93.98 &15.15 & 93.57 &3.64 & 93.28 \\ \hline \hline
    DenseNet (k=12) \cite{huang2017densely} &100.0 & 94.76 &58.01 & 94.51 &36.34 & 94.29 &13.31 & 94.08 &4.29 & 93.57  \\\hline
    ResNext50 $(32\times4d)$ \cite{xie2017aggregated}&100.0 & 96.03 &62.10 & 95.95 &35.60 & 95.84 &16.09 & 95.69 &4.76 & 95.47  \\ \hline\hline
    EfficientNet-B0 \cite{tan2019efficientnet}&100.0 & 98.20 &64.11 & 98.13 &35.94 & 98.01 &18.88 & 97.84 &5.43 & 97.57  \\\hline
    EfficientNet-B2 \cite{tan2019efficientnet}&100.0 & 98.41 &65.16 & 98.35 &37.87 & 98.23 &18.13 & 98.02 &4.69 & 97.88 \\\hline
    EfficientNet-B4 \cite{tan2019efficientnet}&100.0 & 98.70 &64.43 & 98.59 &36.27 & 98.47 &16.40 & 98.23 &4.91 & 98.14 \\\hline
    
    \end{tabular}}
    \label{tab:size_comparison}
\end{table}

\subsection{Intermediate Knowledge Distillation}
 The intermediate block knowledge (feature map representation) is additionally distilled onto every compressed student from the pseudo teacher. This provides a more stricter training and distillation regime for all the compressed students. The loss in every student's representational capacity due to compression is countered, by making each student block try and learn the intermediate feature map of its corresponding pseudo teacher block. The feature map pairs are compared using traditional Mean Squared Error loss, on which the network is trained to reduce any differences between them. Since the number of feature map channels varies across every corresponding student and pseudo teacher block, an adaptation layer consisting of pointwise convolution (1$\times$1 kernel) is used to map compressed student block channels to its pseudo teacher counterpart. Figure \ref{fig:training_time_comparison} (a) presents this idea in detail for an EfficientNet-B0 \cite{tan2019efficientnet} based ensemble. The intermediate knowledge is transferred at three locations corresponding to the three blocks in every student branch as shown in Figure \ref{fig:block_diagram}.

\subsection{Knowledge Distillation Based Training}
The overall ensemble is trained using a combination of three separate losses which are described in detail as follows: \par
\textbf{Cross-Entropy Loss} Each student model is trained individually on classical cross entropy loss [Equation~\ref{eq1},~\ref{eq2}] with one hot vector of labels and student output logits. This loss helps each student directly train on a given dataset. This loss procedure makes the pseudo teacher learn alongside the compressed students as the framework doesn't use pretrained weights of any sort. It also helps the ensemble teacher gain richer knowledge of the dataset as it incorporates combination of every student's learnt knowledge. It additionally enables the framework to avoid training the ensemble teacher separately and is trained implicitly through the students. The output softmax distribution and combined normal loss can be expressed as follows,

\begin{equation} \label{eq1}
    X_{ijk} = \frac{exp(x_{ijk})}{\sum_{k=1}^{C} exp(x_{ijk})}
\end{equation}

\begin{equation} \label{eq2}
    L^{Normal} = \sum_{i=1}^{S} \sum_{j=1}^{N} \sum_{k=1}^{C} - \mathbbm{1}_{jk}\log(X_{ijk})
\end{equation}
where $i,j,k$ represents student, batch sample and class number indices respectively. $\mathbbm{1}_{jk}$ is an one hot label indicator function for $j^{th}$ sample and $k^{th}$ class. Similarly, $x_{ijk}$ is a single output logit from $i^{th}$ student model for $j^{th}$ batch sample and $k^{th}$ class and $X_{ijk}$ is its corresponding softmax output. \par   

\textbf{Intermediate Loss} For every pseudo teacher and compressed student pair, the output feature maps from every pseudo teacher block are compared to the ones at its corresponding compressed student block. In order to facilitate an effective knowledge distillation between these respective map pairs, the compressed student maps are first passed through an adaptation layer which as mentioned earlier is a simple $1 \times1$ convolution, mapping the student map channels to the pseudo teacher map channels. A Mean Squared Error loss is used to compare each single element of a given pseudo teacher-student feature map pair. This loss is averaged across the batch. The loss for a single block across all students can be expressed as follows,
\begin{equation}
    l^{intermediate}_{block} = \sum_{l=2}^{S} \bigg (  \sum_{m=1}^{N} (|x_{m}^{PT} - x_{m}^{l}|)^2 \bigg )
\end{equation}
where $x_{m}^{l}$ is a feature map of size $H \times W \times C$ corresponding to $m^{th}$ batch sample of the $l^{th}$ student model. $|.|^2$ represents element wise squared L2 norm. $l=1$ represents the pseudo teacher, also designated as PT in $x_{m}^{PT}$ which is the corresponding pseudo teacher feature map. The overall intermediate loss can be expressed as: 
\begin{equation}
    L^{intermediate} = \sum_{b=1}^{B} l^{intermediate}_{b}
\end{equation}
This loss is used to update only the compressed student model parameters in order to have the compressed student learn from the pseudo teacher and not the other way round. In our experiments we observed that the mean of adaptation layer weights is on average lower for larger student models. This in turn propagates a smaller model response term in the intermediate loss equation, thus increasing their losses slightly compared to thinner students. This helps balance this loss term across all students. \par

\textbf{Knowledge Distillation Loss} To facilitate global knowledge transfer from the ensemble teacher to each of the students, a KD loss in form of Kullback-Leibler Divergence Loss is incorporated between the ensemble teacher and student outputs. The outputs of the ensemble teacher and each respective student are softened using a certain temperature $T$ to help students learn efficiently from highly confident ensemble teacher predictions where the wrong class outputs are almost zero. The softened softmax and overall KD loss can be expressed as follows,

\begin{equation} \label{softened_softmax}
    X_{ijk} = \frac{exp(\frac{x_{ijk}}{T})}{\sum_{k=1}^{C} exp(\frac{x_{ijk}}{T})}
\end{equation}
\begin{equation}
    L^{KD} = \sum_{i=1}^{S}  \sum_{j=1}^{N} \sum_{k=1}^{C} X_{jk}^{T} \log \Big ( \frac{X_{jk}^{T}}{X_{ijk}} \Big ) 
\end{equation}
where $X_{ijk}$ is the softened softmax output of the $i^{th}$ student for $j^{th}$ batch sample and $k^{th}$ class. Similarly $X_{jk}^{T}$ represents the ensemble teacher softened softmax output for $j^{th}$ batch sample and $k^{th}$ class.\par

\textbf{Combined Loss} The above presented three losses are combined using a weighted combination, on which the entire framework is trained to reduce this overall loss. This can be mathematically expressed as,
\begin{equation} \label{combination_equation}
    L = \alpha L^{Normal} + \beta L^{intermediate} + \gamma L^{KD}
\end{equation}
 The optimal weight value combination which was found out to be $\alpha = 0.7 $, $\beta =0.15 $, $\gamma =0.15 $ is discussed in detail in an ablation study presented in later sections. 

\section{Experiments}

\textbf{Datasets.} We incorporate four major academic datasets: (1) CIFAR10 dataset \cite{krizhevsky2009learning} which contains 50,000/10,000 training/test samples drawn from 10 classes. Each class has 6,000 images included in both training and test set sized at $32 \times 32$ pixels. (2) CIFAR100 dataset \cite{krizhevsky2009learning} which contains 50,000/10,000 training/test samples drawn from 100 classes. Each class has 600 images included in both training and test set sized at $32 \times 32$ pixels. (3) SVHN dataset \cite{netzer2011reading} which contains 73,257/26,032 training/test samples drawn from 10 classes. Each class represents a digit from 0 to 9. Each image is sized at $32 \times 32$ pixels. (4) ImageNet dataset \cite{deng2009imagenet} is a comprehensive database containing around 1.2 million images, specifically 1,281,184/50,000 training/testing images drawn from 1000 classes.\\

\noindent{\textbf{Experimental Hypothesis.}} Experiments are conducted in order to: (1) compare every compressed student's test set performance trained using our ensemble framework versus simply training each one of them individually without any knowledge distillation component. These experiments help validate the advantages of using an ensemble teacher and intermediate knowledge transfer for every compressed student compared to using only the traditional individual cross entropy loss based training. (2) Compare the test set accuracy of our ensemble teacher to other notable ensemble knowledge distillation based techniques in literature on all four mentioned datasets to prove our framework's overall superiority and effectiveness, which are presented in Table \ref{tab:method_comparison}. (3) Compare the time taken for training our five student based ensemble versus the combined time taken for training each of those students individually. This comparison helps substantiate the training time benefits of our hybrid multi-student architecture compared to training each student alone either sequentially or in parallel. These are presented in Figure \ref{fig:training_time_comparison} (b).\par     

\begin{table}[t]
\begin{center}
\caption{Individual Test Set performance comparison for five compressed students trained using our ensemble and using baseline training on CIFAR10 dataset. Reported results are averaged over five individual experimental runs.}
\resizebox{.98\textwidth}{!}{%
\begin{tabular}{c| c| c |c |c| c |c |c |c |c |c}
\hline
\bf Classification & \multicolumn{10}{c}{\bf Student Test Accuracy (\%)}   \\\cline{2-11}
\bf    Model   & \multicolumn{2}{c}{First} & \multicolumn{2}{c}{Second}  & \multicolumn{2}{c}{Third}& \multicolumn{2}{c}{Fourth}& \multicolumn{2}{c}{Fifth} \\\cline{2-11}
& Baseline & Ensemble & Baseline & Ensemble & Baseline & Ensemble & Baseline & Ensemble & Baseline & Ensemble \\\hline

Resnet20 \cite{he2016deep}&91.34 & \bf 92.13 &91.12 &\bf92.18 &90.89 &\bf91.78 &90.16 &\bf91.45 &89.67 &\bf91.03 \\\hline
Resnet32 \cite{he2016deep}&92.12 &\bf92.95 &91.94 &\bf92.76 &91.56 &\bf92.45&91.07 &\bf92.11 &90.47 &\bf91.78 \\\hline
Resnet44 \cite{he2016deep}&92.94 &\bf93.45 &92.67 &\bf93.29 &92.24 &\bf93.11 &91.97 &\bf92.89 &91.23 &\bf92.56 \\\hline
Resnet110 \cite{he2016deep}&93.51 &\bf94.24 &93.25 &\bf94.18 &93.11 &\bf93.98 &92.86 &\bf93.57 &92.27 &\bf93.28 \\ \hline\hline
Densenet-BC (k=12)\cite{huang2017densely} &94.02 &\bf94.76 &93.78 &\bf94.51 &93.52 &\bf94.29 &93.24 &\bf94.08 &92.85 &\bf93.57 \\\hline
ResNext50 $(32\times4d)$\cite{xie2017aggregated} &95.78 &\bf96.03 &95.56 &\bf95.95 &95.27 &\bf95.84 &95.09 &\bf95.69 &94.97 &\bf95.47 \\ \hline\hline
EfficientNet-B0\cite{tan2019efficientnet} &97.82 &\bf98.20 &97.58 &\bf98.13 &97.28 &\bf98.01 &97.04 &\bf97.84 &96.73 &\bf97.57 \\\hline
EfficientNet-B2\cite{tan2019efficientnet} &98.21 &\bf98.41 &98.13 &\bf98.35 &97.99 &\bf98.23 &97.77 &\bf98.02 &97.41 &\bf97.88 \\\hline
EfficientNet-B4 \cite{tan2019efficientnet}&98.56 &\bf98.70 &98.36 &\bf98.59 &98.21 &\bf98.47 &98.04 &\bf98.23 &97.92 &\bf98.14 \\\hline

\end{tabular}}
\label{tab:cifar10_exp_1}
\end{center}
\end{table}

\begin{table}[t]
\begin{center}
\caption{Individual Test Set performance comparison for five compressed students trained using our ensemble and using baseline training on CIFAR100. Reported results are averaged over five individual experimental runs.}
\resizebox{.98\textwidth}{!}{%
\begin{tabular}{c| c| c |c |c| c |c |c |c |c |c}
\hline
\bf Classification & \multicolumn{10}{c}{\bf Student Test Accuracy (\%)}   \\\cline{2-11}
\bf    Model   & \multicolumn{2}{c}{First} & \multicolumn{2}{c}{Second}  & \multicolumn{2}{c}{Third}& \multicolumn{2}{c}{Fourth}& \multicolumn{2}{c}{Fifth} \\\cline{2-11}
& Baseline & Ensemble & Baseline & Ensemble & Baseline & Ensemble & Baseline & Ensemble & Baseline & Ensemble \\\hline

Resnet32 \cite{he2016deep}&70.21 & \bf 70.97 &67.87 &\bf 68.24 & 64.17 &\bf 65.67 & \bf 61.85 & 61.17 & 39.12 &\bf 42.17 \\\hline
Resnet44 \cite{he2016deep}&71.12 &\bf 71.76 & 68.42 &\bf 69.12 & 65.69 &\bf 67.04 & 62.31 &\bf 62.87 & 40.82 &\bf 43.11 \\\hline
Resnet56 \cite{he2016deep}& 71.59 &\bf 72.16 &\bf 68.45 & 68.39 & 65.37 &\bf 66.21 &\bf 62.42 & 62.21 & 41.19 &\bf 43.27 \\\hline
Resnet110 \cite{he2016deep}&72.64 &\bf 72.81 &69.53 &\bf 70.14 &67.12 &\bf67.73 &64.58 &\bf65.08 &42.26 &\bf 46.76 \\ \hline\hline
Densenet-BC (k=12)\cite{huang2017densely} &75.79 &\bf 75.96 & 71.97 &\bf 72.39 &\bf 70.23 & 70.09 & 67.13 &\bf 68.14 & 45.41 &\bf 49.12 \\\hline
ResNeXt50 $(32\times4d)$\cite{xie2017aggregated} &72.37 &\bf72.59 &70.19 &\bf70.32 &67.02 &\bf67.81 &65.19 &\bf65.72 & 42.82 &\bf 45.29 \\ \hline\hline
EfficientNet-B0\cite{tan2019efficientnet} &87.17 &\bf88.12&85.78 & \bf 86.94 & 83.25& \bf 85.14& 80.24 &\bf 83.21&76.35 &\bf 78.45 \\\hline
EfficientNet-B2\cite{tan2019efficientnet} &89.05 &\bf 89.31 &87.34 & \bf 88.78 &85.23 &\bf87.58 &82.14& \bf 84.13&79.34 & \bf 81.12 \\\hline
EfficientNet-B4 \cite{tan2019efficientnet}&90.26 &\bf90.81 &88.59 &\bf89.78 &86.34 &\bf 88.04& 84.32& \bf 86.78& 81.34&\bf 84.10 \\\hline

\end{tabular}}
\label{tab:cifar100_exp_1}
\end{center}
\end{table}

\begin{table}[t]
\begin{center}
\caption{Comparison of notable knowledge distillation and ensemble based techniques with our ensemble teacher reported test accuracy performance (Error rate \%). The best performing model accuracy is chosen for DML.}
\resizebox{0.95\textwidth}{!}{%
\begin{tabular}{c|c|c|c|c|c|c|c|c}
\hline
\bf  Ensemble    & \multicolumn{8}{c}{\bf Dataset}   \\\cline{2-9}
\bf   Technique  & \multicolumn{2}{c}{CIFAR10} & \multicolumn{2}{c}{CIFAR100} & \multicolumn{2}{c}{SVHN} & \multicolumn{2}{c}{ImageNet}   \\\cline{2-9}
& ResNet-32 & ResNet-110& ResNet-32 & ResNet-110& ResNet-32 & ResNet-110& Resnet-18 & ResNeXt-50 \\ \hline
KD-ONE \cite{zhu2018knowledge}&5.99 &5.17 &26.61 &21.62 & \bf 1.83 & 1.76 & 29.45 & 21.85\\\hline
DML \cite{zhang2018deep}& --& -- & 29.03 & 24.10 & --  & -- & -- & --  \\\hline
Snopshot Ensemble \cite{huang2017snapshot}& -- & 5.32 & 27.12 & 24.19 & -- & 1.63 & -- & -- \\\hline
Ours & \bf 5.73 & \bf 4.85 & \bf 26.09 & \bf 21.14 &  1.97 & \bf 1.61 & \bf 29.34 & \bf 21.17 \\\hline
\end{tabular}}
\label{tab:method_comparison}
\end{center}
\end{table}

\begin{table}[t]
\begin{center}
\caption{Individual Test Set performance comparison for five compressed students trained using our ensemble and using baseline training on SVHN. Reported results are averaged over five individual experimental runs.}
\resizebox{0.98\textwidth}{!}{%
\begin{tabular}{c| c| c |c |c| c |c |c |c |c |c}
\hline
\bf Classification & \multicolumn{10}{c}{\bf Student Test Accuracy (\%)}   \\\cline{2-11}
\bf    Model   & \multicolumn{2}{c}{First} & \multicolumn{2}{c}{Second}  & \multicolumn{2}{c}{Third}& \multicolumn{2}{c}{Fourth}& \multicolumn{2}{c}{Fifth} \\\cline{2-11}
& Baseline & Ensemble & Baseline & Ensemble & Baseline & Ensemble & Baseline & Ensemble & Baseline & Ensemble \\\hline

Resnet20 \cite{he2016deep}&96.64 & \bf 97.10 &95.03 &\bf96.92 &94.45 &\bf95.53 & \bf 92.12 & 92.03 &89.58 &\bf92.67 \\\hline
Resnet32 \cite{he2016deep}&96.78 &\bf96.92 &95.67 &\bf96.31 & \bf 94.85 & 94.61 & 92.78 &\bf95.03 &90.75 &\bf92.89 \\\hline
Resnet44 \cite{he2016deep}&97.23 &\bf97.46 & \bf 96.38 & 96.26 &95.35 &\bf96.32 &93.26 &\bf95.76 &91.24 &\bf93.47 \\\hline
Resnet110 \cite{he2016deep}&97.64 &\bf97.87 &96.61 &\bf97.84 &95.83 &\bf96.81 &93.73 &\bf95.90 &91.77 &\bf93.78 \\ \hline\hline
Densenet-BC (k=12)\cite{huang2017densely} &97.92 &\bf98.03 &97.31 &\bf98.02 &96.12 &\bf97.59 & \bf 94.58 & 94.25 &92.15 &\bf94.17 \\\hline
ResNext50 $(32\times4d)$\cite{xie2017aggregated} &97.65 &\bf97.88 & \bf 96.84 & 96.69 &95.72 &\bf96.64 & \bf 94.79 & 94.23 &91.73 &\bf93.80 \\ \hline\hline
EfficientNet-B0\cite{tan2019efficientnet} &97.53 &\bf97.72 &97.07 &\bf97.79 &95.52 &\bf96.71 &\bf 94.44 & 94.26 &91.12 &\bf93.34 \\\hline
EfficientNet-B2\cite{tan2019efficientnet} &97.75 &\bf97.92 & \bf 97.76 & 97.63 &95.87 &\bf96.92 &93.37 &\bf96.24 & \bf 91.42 & 91.29 \\\hline
EfficientNet-B4 \cite{tan2019efficientnet}&98.16 &\bf98.56 &97.79 &\bf98.03 & \bf 96.71 & 96.48 &93.64 &\bf96.83 &91.75 &\bf94.17 \\\hline

\end{tabular}}
\label{tab:SVHN_exp_1}
\end{center}
\end{table}

\noindent{\textbf{Performance Metrics.}} We compare the test set accuracy (Top-1) of each of our student models within the ensemble, trained using our framework and as an individual baseline model with only the traditional cross entropy loss. For each of our ensemble students, this test set accuracy is computed as an average of the best student test accuracies achieved during each of five conducted runs.\par

\noindent{\textbf{Experimental Setup.}} For fair comparison, we keep the training schedule the same for both our ensemble framework and baseline training. Specifically, for ResNet, DenseNet and ResNeXt models SGD is used with Nesterov momentum set to 0.9, following a standard learning rate schedule that drops from 0.1 to 0.01 at 50\% training and to 0.001 at 75\%. For EfficientNet models RMSProp optimizer is implemented with decay 0.9 and momentum 0.9 and initial learning rate of 0.256 that decays by 0.97 every 3 epochs. The models are trained for 350/450/50/100 epochs each for the CIFAR10/CIFAR100/SVHN/ImageNet datasets respectively.

\begin{table}[t]
\begin{center}
\caption{Individual Test Set performance comparison for five compressed students trained using our ensemble and using baseline training on ImageNet (Top-1 accuracy). Reported results are averaged over five individual experimental runs.}
\resizebox{0.98\textwidth}{!}{%
\begin{tabular}{c| c| c |c |c| c |c |c |c |c |c}
\hline
\bf Classification & \multicolumn{10}{c}{\bf Student Test Accuracy (Top-1 accuracy \%)}   \\\cline{2-11}
\bf    Model   & \multicolumn{2}{c}{First} & \multicolumn{2}{c}{Second}  & \multicolumn{2}{c}{Third}& \multicolumn{2}{c}{Fourth}& \multicolumn{2}{c}{Fifth} \\\cline{2-11}
& Baseline & Ensemble & Baseline & Ensemble & Baseline & Ensemble & Baseline & Ensemble & Baseline & Ensemble \\\hline

Resnet18 \cite{he2016deep}&69.73 & \bf 70.47 & 67.27 &\bf 67.61 & 62.98 &\bf 64.88 & 59.47 &\bf61.17 &55.23 &\bf58.52 \\\hline
Resnet34 \cite{he2016deep}&73.22 &\bf74.13 &71.95 &\bf73.64 &67.62 &\bf 69.32 & 63.07 &\bf64.19 &60.76 &\bf61.29 \\\hline
Resnet50 \cite{he2016deep}&76.18 &\bf76.52 & \bf 75.43 & 75.32 &70.16  &\bf 71.93 & 66.89  &\bf 69.46 &62.24 &\bf66.78 \\\hline
Resnet101 \cite{he2016deep}&77.31 &\bf77.97 &76.27 &\bf76.71 &73.49 &\bf74.04 &69.47 &\bf71.10 &65.79 &\bf68.57 \\ \hline\hline
Densenet-121\cite{huang2017densely} &74.96 &\bf75.82 &73.94 &\bf74.17 & \bf 68.53 & 68.44 & 66.64 &\bf67.83 &63.42 &\bf66.09 \\\hline
ResNext50 $(32\times4d)$\cite{xie2017aggregated} &77.58 &\bf78.19 &76.62 &\bf77.85 & \bf 73.45  & 73.37 & 69.73  &\bf 70.89 &65.82 &\bf 68.48 \\ \hline

\end{tabular}}
\label{tab:Imagenet_exp_1}
\end{center}
\end{table}

\subsection{Evaluation of our online model compression framework}
\begin{figure}[t]
    \centering
    \includegraphics[width=\textwidth]{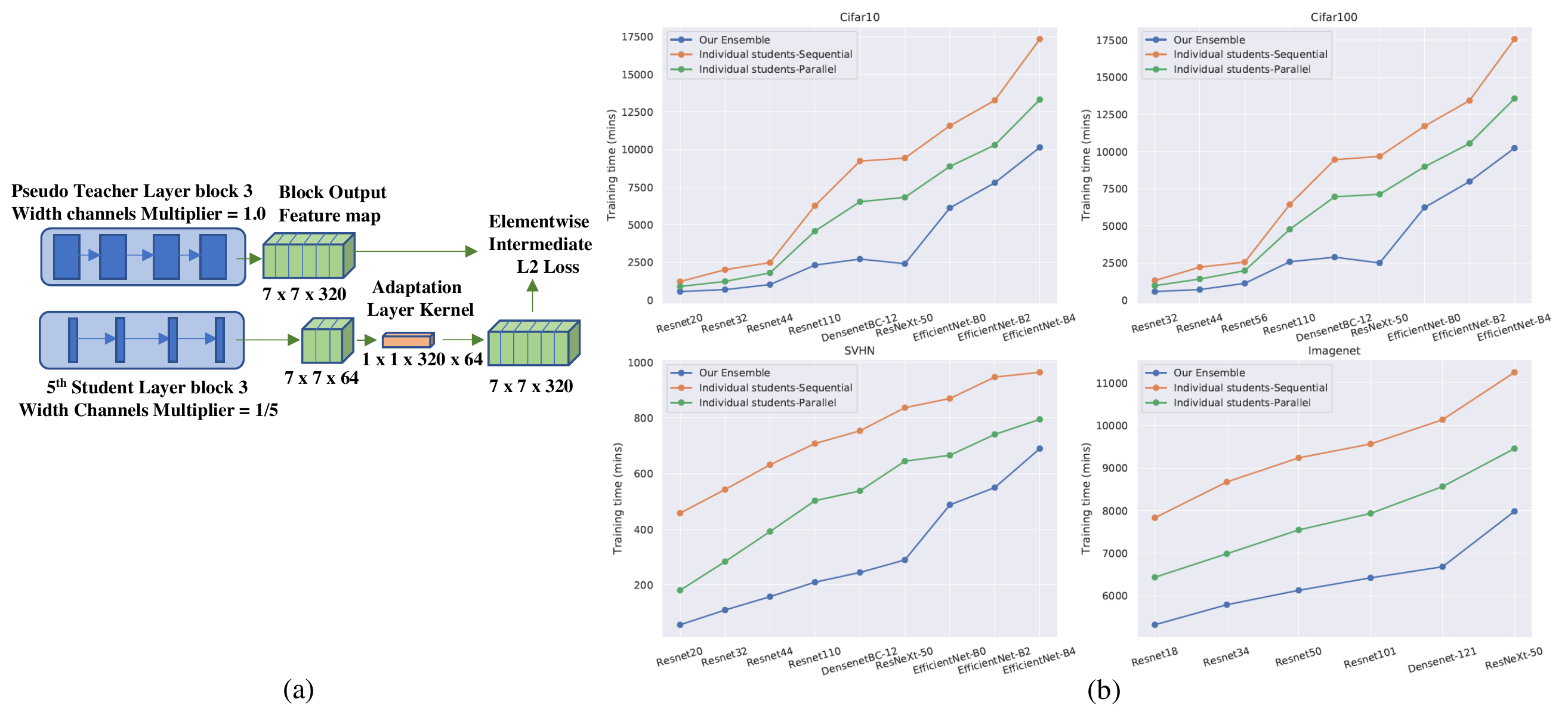}
    \caption{(a) \textit{Channel Adaptation Logic} for mapping output channels of an \textit{EfficientNet-B0} model based ensemble, depicting Block 3 outputs of the pseudo Teacher and 5th student in the ensemble. (b) Comparison of our ensemble framework training time (Blue) to the combined training time of individual baseline students performed sequentially (Orange) and in parallel (Green). Timings recorded for training carried out on a GPU cluster of Nvidia GTX 1080Ti.}
    \label{fig:training_time_comparison}
\end{figure}

\textbf{Results on CIFAR10 and CIFAR100.} Tables \ref{tab:cifar10_exp_1},\ref{tab:cifar100_exp_1} present our experimental results for CIFAR10 and CIFAR100 dataset respectively. Each compressed student's test set performance is on an average 1\% better using our ensemble framework as compared to the simple baseline training for both the datasets. Our ensemble teacher also provides the best Test set accuracy when compared to the teacher accuracies of three other ensemble knowledge distillation techniques for ResNet32 and ResNet110 models as presented in Table \ref{tab:method_comparison}. Our framework provides substantial training time benefits for all models tested with CIFAR10 and CIFAR100 datasets as presented in Fig \ref{fig:training_time_comparison} (b). For fair comparison the ensemble and each of the baseline students are trained for the same number epochs on both datasets. Notably, training a five student ensemble of an EfficientNet-B4 architecture is roughly 7.5K GPU minutes quicker as compared to their combined individual baseline training.

\noindent{\textbf{Results on SVHN and ImageNet.}} Tables \ref{tab:SVHN_exp_1},\ref{tab:Imagenet_exp_1} present our experimental results for SVHN and ImageNet datasets respectively. Again, each compressed student test set performance is on an average 1\% better using our ensemble framework as compared to the simple baseline training for both the datasets. Notably for both datasets, the heavily compressed fourth and fifth students perform around 3\% on average better than their baseline counterparts. This provide an excellent evidence of our framework's efficient knowledge transfer capabilities for the heavily compressed student cases. Similar to the aforementioned datasets, our ensemble teacher provides the best Test set accuracy when compared to the ensemble teacher accuracy of three other ensemble knowledge distillation based techniques for ResNet32 and ResNet110 models as presented in Table \ref{tab:method_comparison}. 

\section{Ablation Studies}

\textbf{Loss Contribution Ratios.} A selective grid search was conducted for the optimal values of loss contribution ratios, specifically $\alpha,\beta,\gamma$ referenced in Equation \ref{combination_equation}. The grid search was conducted with the constraint that the ratio should sum to one which would represent the overall loss factor. The study was carried out using ResNet110 \cite{he2016deep} model on the CIFAR100 \cite{krizhevsky2009learning} dataset. Firstly, a grid search was conducted for $\alpha$ which is the normal loss contribution ratio. The other two contribution ratio namely $\beta , \gamma$ were kept equal to half the fraction left from subtracting the grid search $\alpha$ from 1. This study is presented in Table \ref{tab:alpha_study}. All accuracies are averaged over five runs to reduce any weight initialization effect. The value of 0.7 provided the best weighted average accuracy across all the students. Weights of $\frac{1}{5},\frac{2}{5},\frac{3}{5},\frac{4}{5},1$ were assigned to the students, with higher importance given to accuracy achieved by more compressed student.\par

With the value of $\alpha$ set as 0.7, a grid search was then conducted for the value of $\beta$ and $\gamma$. This study is presented in Table \ref{tab:beta_gamma_study}. Here also the same weighted average technique was used to find the effective student test accuracy. $\beta = 0.15$ and $\gamma = 0.15$ gave the optimal performance and were thus selected as the final contribution ratios for our framework. These final ratios seem to indicate the major importance of cross entropy loss with $\alpha = 0.7$ in individual student's training and equal importance of intermediate and output distribution knowledge transfer with $\beta = \gamma = 0.15$ for the knowledge distillation process.\par       

\begin{table}[t]
\begin{center}
\caption{Ablation Study for $\alpha$ contribution ratio grid search conducted with ResNet110 \cite{he2016deep} model on CIFAR100 \cite{krizhevsky2009learning} dataset. Weighted average technique used for calculating effective student model accuracy with higher weight given to more compressed students.}
\resizebox{0.75\textwidth}{!}{%
\begin{tabular}{c c c c c c c c}
\hline
\bf $\alpha$ & \multicolumn{5}{c}{\bf Student Test Accuracy (\%)}  &\bf Ensemble Teacher &\bf Weighted  \\\cline{2-6}
             & \bf 1 & \bf  2& \bf  3& \bf  4& \bf  5 & \bf Test Accuracy (\%) & \bf Average \\\hline
0.3 & 69.58 & 64.47 & 63.21 & 55.64 & 39.57 & 69.08 & 161.71 \\ \hline
0.4 & 68.27 & 65.36 & 62.35 & 57.54 & 40.14 & 68.81 & 163.38 \\ \hline
0.5 & 71.32 & 69.58 & 67.92 & 62.69 & 45.16 & 72.15 & 178.16 \\ \hline
0.6 & 72.11 & 67.82 & 65.55 & 60.93 & 43.19 & 71.64 & 172.81 \\ \hline
\bf0.7 & \bf71.25 & \bf69.32 & \bf67.29 & \bf62.16 & \bf47.23 & \bf72.51 &\bf 179.31 \\ \hline
0.8 & 71.05 & 70.21 & 68.01 & 62.61 & 45.10 & 71.85 & 178.29 \\ \hline
0.9 & 69.21 & 66.56 & 63.12 & 58.85 & 45.76 & 71.86 & 171.18 \\ \hline
\end{tabular}}
\label{tab:alpha_study}
\end{center}
\end{table}

\noindent{\textbf{Knowledge distillation Temperature.}} A temperature variable $T$ is used to soften the student and ensemble teacher model logits before computing its respective softmax distribution as referenced in Equation \ref{softened_softmax}. A grid search was conducted for its optimal value, which would facilitate optimum knowledge transfer from the ensemble teacher to each student model. Similar to the previous study, we incorporate a ResNet110 \cite{he2016deep} model to train on CIFAR100 \cite{krizhevsky2009learning} dataset. This study is presented in Table \ref{tab:temperature_study}. The resulting optimal value of 2 is used for all of our conducted experiments. The study results provide evidence to the fact that higher temperature values tend to over-soften the output logits leading to sub optimal knowledge transfer and test accuracy gains.   

\begin{table}[t]
\begin{center}
\caption{Ablation Study for $\beta , \gamma$ contribution ratios grid search conducted with ResNet110 \cite{he2016deep} model on CIFAR100 \cite{krizhevsky2009learning} dataset. Weighted average technique used for calculating combined student model test accuracy with higher weight given to more compressed students. $\alpha$ is set at optimal value of 0.7 referred from Table \ref{tab:alpha_study}. }
\resizebox{0.85\textwidth}{!}{%
\begin{tabular}{c c c c c c c c c}
\hline
\bf $\beta$ & \bf $\gamma$ & \multicolumn{5}{c}{\bf Student Test Accuracy (\%)}  &\bf Ensemble Teacher &\bf Weighted  \\\cline{3-7}
        &     & \bf 1 & \bf  2& \bf 3 & \bf  4& \bf  5 & \bf Test Accuracy (\%) & \bf Average \\\hline
0.05 & 0.25 & 68.57 & 66.69 & 64.34 & 59.92 & 47.26 & 71.97 & 174.19 \\ \hline
0.1 & 0.2 & 68.72 & 68.19 & 65.94 & 62.09 & 44.88 & 71.73 & 175.14 \\ \hline
\bf 0.15 &\bf 0.15&\bf 69.37 &\bf 68.39 &\bf 66.44 &\bf 62.17 &\bf 46.82 &\bf 72.28 &\bf 177.65 \\ \hline
0.2 & 0.1& 66.79 & 64.92 & 62.46 & 57.78 & 41.34 & 69.64 & 164.37 \\ \hline
0.25 & 0.05 & 67.97 & 67.22 & 65.37 & 61.08 & 46.69 & 71.12 & 175.26 \\ \hline
\end{tabular}}
\label{tab:beta_gamma_study}
\end{center}
\end{table}

\begin{table}[t]
\begin{center}
\caption{Ablation Study for softmax temperature (T) grid search conducted with ResNet110 \cite{he2016deep} model on CIFAR100 \cite{krizhevsky2009learning} dataset. Mean accuracy computed using only student test accuracies.}
\resizebox{0.85\textwidth}{!}{%
\begin{tabular}{c c c c c c c c}
\hline
\bf Temperature & \multicolumn{5}{c}{\bf Student Test Accuracy (\%)}  &\bf Ensemble Teacher &\bf Mean  \\\cline{2-6}
     \bf T        & \bf 1 & \bf  2& \bf  3& \bf  4& \bf  5 & \bf Test Accuracy (\%) & \bf Accuracy (\%) \\\hline
1 & 70.29 & 67.16 & 64.22 & 62.46 & 44.75 & 71.57 & 61.78 \\ \hline
\bf 2 &\bf 71.89 &\bf 69.56 &\bf 68.43 &\bf 59.48 &\bf 47.26 &\bf 71.02 &\bf 63.32 \\ \hline
3 & 69.14 & 68.18 & 65.33 & 57.52 & 46.33 & 69.36 & 61.3 \\ \hline
4 & 68.35 & 66.77 & 64.36 & 56.41 & 46.45 & 69.37 & 60.49 \\ \hline
5 & 66.58 & 66.95 & 65.67 & 56.1 & 42.62 & 69.27 & 59.58 \\ \hline
6 & 66.92 & 66.28 & 65.33 & 55.97 & 43.06 & 69.34 & 59.51 \\ \hline
\end{tabular}}
\label{tab:temperature_study}
\end{center}
\end{table}

\section{Discussion}
The performed experiments provide a strong evidence of the efficient compression and generalization capabilities of our framework over individual baseline training for every compressed student model. In most of the experiments the ensemble teacher's test accuracy is much better than any of its ensemble students and their baseline counterparts. This additional test accuracy gain can be attributed to the joint ensemble knowledge learnt by the framework.\par
\begin{figure}[t]
    \centering
    \includegraphics[width= 1.0\linewidth,height=0.2\textheight]{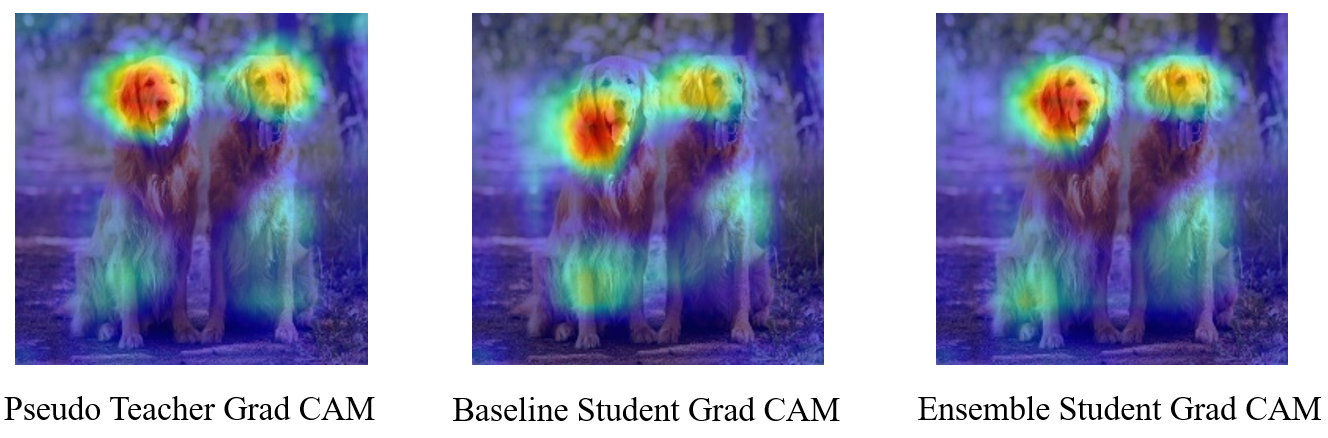}
    \caption{Gradient Class Activation Mapping (Grad CAM) \cite{selvaraju2017grad} comparison of a EfficientNet-B4 based ensemble pseudo teacher and one of its compressed students with that of its respective individually trained student. The ensemble student's CAM is more accurate compared to that of baseline student. Also the former follows the pseudo teacher more closely as compared to the latter, which provides evidence of the effective knowledge distillation taking place in our ensemble framework.}
    \label{fig:grad_cam_comparison}
\end{figure}

The intermediate knowledge transfer from the pseudo teacher onto each one of the compressed students helps guide every student compressed block to reproduce the same transformations its respective higher capacity pseudo teacher block is learning. Enabling the low capacity compressed block to try and imitate the higher capacity pseudo teacher block helps reduce any redundant (sub-optimal) transformations inside the student block that would generally be present during baseline training. This is substantiated by the fact that the test accuracy gains of heavily compressed students, specifically the fourth and fifth students in the ensemble are substantial over their baseline counterparts. Figure \ref{fig:grad_cam_comparison} presents a comparison of the gradient based class activation mapping (Grad-CAM) \cite{selvaraju2017grad}  of the last block of an EfficientNet-B4 framework pseudo teacher and one of its compressed students. These are compared to the Grad-CAM of the same compressed student with baseline training. The smaller differences between the Grad-CAMs of pseudo teacher and its ensemble student compared to those between the pseudo teacher and the baseline student provide evidence of how our efficient knowledge distillation helps the student imitate the pseudo teacher and learn better as compared to the baseline student.

\section{Conclusion}
We present a novel model compression technique using an ensemble knowledge distillation learning procedure without requiring the need of any pretrained weights. The framework manages to provide multiple compressed versions of a given base (pseudo teacher) model simultaneously, providing gains in each of the participating model's test performance and in overall framework's training time compared to each model's individual baseline training. Comprehensive experiments conducted using a variety of current state-of-the-art image classification based models and benchmarked academic datasets provide substantial evidence of the framework's effectiveness. It also provides an account of the highly modular nature of the framework which makes it easier to incorporate any existing classification model into the framework without any major modifications. It manages to provide multiple efficient versions of the same, compressed to varying degree without making any major manual architecture changes on the user's part.     

%
%
\bibliographystyle{splncs04}
\bibliography{egbib}
\end{document}